\pdfoutput=1

\documentclass{article}
\usepackage{ijcai21}

\usepackage{times}
\usepackage{soul}
\usepackage{url}
\usepackage[hidelinks]{hyperref}
\usepackage[utf8]{inputenc}
\usepackage[small]{caption}
\usepackage{graphicx}
\usepackage{amsmath}
\usepackage{amsthm}
\usepackage{booktabs}
\usepackage{multirow}
\usepackage{subfigure}
\usepackage{algorithm}
\usepackage{algorithmic}
\usepackage{listings}
\usepackage{color}
\definecolor{dkgreen}{RGB}{84,146,181}
\definecolor{gray}{rgb}{0.5,0.5,0.5}
\definecolor{mauve}{rgb}{0.58,0,0.82}
\DeclareCaptionFormat{mylst}{\hrule#1#2#3}
\title{Towards Compact Single Image Super-Resolution via Contrastive Self-distillation}

\author{
Yanbo Wang$^1$
\and
Shaohui Lin$^{1*}$\and
Yanyun Qu$^2$\and\\
Haiyan Wu$^1$\and
Zhizhong Zhang $^1$\and
Yuan Xie $^{1*}$\And
Angela Yao $^3$
\affiliations
$^1$East China Normal University\\
$^2$Xiamen University\\
$^3$National University of Singapore\\
\emails
\{51205901021, 51194501183\}@stu.ecnu.edu.cn,
\{shlin,yxie,zzzhang\}@cs.ecnu.edu.cn,
yyqu@xmu.edu.cn,
ayao@comp.nus.edu.sg,
}

\begin{document}

\maketitle
\footnote{$^*$Corresponding author}

\begin{abstract}
   Convolutional neural networks (CNNs) are highly successful for super-resolution (SR) but often require sophisticated architectures with heavy memory cost and computational overhead,  significantly restricts their practical deployments on resource-limited devices. In this paper, we proposed a novel contrastive self-distillation (CSD) framework to simultaneously compress and accelerate various off-the-shelf SR models. In particular, a channel-splitting super-resolution network can first be constructed from a target teacher network as a compact student network. Then, we propose a novel contrastive loss to improve the quality of SR images and PSNR/SSIM via explicit knowledge transfer. Extensive experiments demonstrate that the proposed CSD scheme effectively compresses and accelerates several standard SR models such as EDSR, RCAN and CARN. \emph{Code is available at \href{https://github.com/Booooooooooo/CSD}{https://github.com/Booooooooooo/CSD}.}
\end{abstract}

\section{Introduction}

Single Image Super-Resolution (SISR) aims to reconstruct a high-resolution (HR) image given a low-resolution (LR) image. 
Recently, convolutional neural networks (CNNs) \cite{kim2016deeply,lim2017enhanced} have dominated SR approaches
by directly learning a mapping from LR input to HR output with a deep neural network. 
However, the existing methods \cite{zhang2018image,wang2018esrgan} tend to explore complex and delicate network architectures for recovering edge structures and missing texture details.  These networks consume significant amounts of memory and computational cost and are therefore impractical to apply in 
resource-limited devices such as wearables and IoT.

Several SR model compression methods have been proposed to remove the redundant parameters. 
Recursive models \cite{yang2018drfn,li2019srfbn} share the SR network's major blocks and reduces the model size and number of parameters.  However, the recursive nature still results in time-consuming inference procedures as the model cost up to 15s per 2K image with 4$\times$ SR on one Intel i9-10980XE CPU \cite{li2019srfbn}.
Parameter quantization~\cite{ma2019efficient,li2020pams}
can reduce memory storage by converting the parameters into lower bits. 
However, the computational complexity is still high as the models are rarely fully quantized. For example,~\cite{ma2019efficient} quantize parameters but still use full-precision activations.  Similarly,~\cite{li2020pams} uses additional float scales
and low-bit weights/activations to approximate the original full-precision weights/activations, still requiring float-based convolutional computation. 
\begin{figure}[t]
\centering
\resizebox{.75\linewidth}{!}{
\begin{minipage}[b]{0.303\linewidth}
    {
    \begin{center}
        \quad \\
    \end{center}
    \includegraphics[width=1\linewidth, height=1.8in]{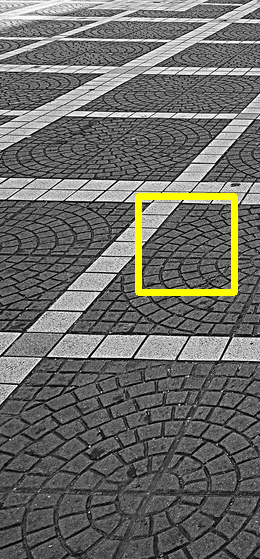}
    \begin{center}
        \small 095 from Urban100\\
        \small \quad
    \end{center}
    }
    \end{minipage}
    \,
    \begin{minipage}[b]{0.27\linewidth}
    {
    \includegraphics[width=1\linewidth, height=0.75in]{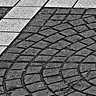}
    \begin{center}
        \small HR\\
        \small PSNR / SSIM
    \end{center}
    \includegraphics[width=1\linewidth, height=0.75in]{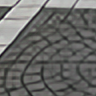}
    \begin{center}
        \small L1 + $L_{percep}$\\
        \small 33.44 dB / 0.3647
    \end{center}
    }
    \end{minipage}
    \,
    \begin{minipage}[b]{0.27\linewidth}
    {
    \includegraphics[width=1\linewidth, height=0.75in]{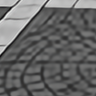}
    \begin{center}
        \small Only L1 loss\\
        \small 33.46 dB / 0.3446
    \end{center}
    \includegraphics[width=1\linewidth, height=0.75in]{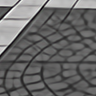}
    \begin{center}
        \small CSD (Ours)\\
        \small \textbf{33.69 dB / 0.4229}
    \end{center}
    }
    \end{minipage}
}
    \caption{Visual comparison of CSSR-Net with different losses.}
    \label{visualLossFig}
\end{figure}

We aim to simultaneously compress and accelerate SR models.  We propose a simple \emph{self-distillation} framework inspired by \cite{yu2018slimmable}, in which a student network is split from a teacher (target) network by using parts of teacher's channels in each layer. We term this student network as \emph{Channel-Splitting Super-Resolution network} (CSSR-Net). 
The teacher and student networks are trained jointly to form two SR models with different computation. 
According to different computation resources 
in devices, we can dynamically allocate these two models, \emph{i.e.} select CSSR-Net if exceeding the required computation overhead in the limited-resource devices, and the teacher model otherwise.

Training CSSR-Net and its teacher network jointly using a reconstruction loss \emph{implicitly} transfers  knowledge from teacher to CSSR-Net. Implicit knowledge transfer considers to transfer the knowledge from the teacher to CSSR-Net by weight sharing and joint training, which only provides limited internal knowledge. This will result in low PSNR/SSIM and low quality HR images on CSSR-Net. Recently, perceptual loss~\cite{johnson2016perceptual} have been widely used as additional knowledge to improve the quality, but the results of CSSR-Net still suffer from blurred lines and edges, as shown in Fig. \ref{visualLossFig}. This is due to limited external  knowledge of the perceptual loss that uses only ground-truth HR images as the upper bound of CSSR-Net and teacher. This begs our rethinking: \emph{Why not select multiple negative samples as the lower bounds to reduce the optimization space and provide more explicit knowledge to improve the student's performance?}

\begin{figure*}[t]
\centering 
\includegraphics[width= 0.9\textwidth, height=2.4in]{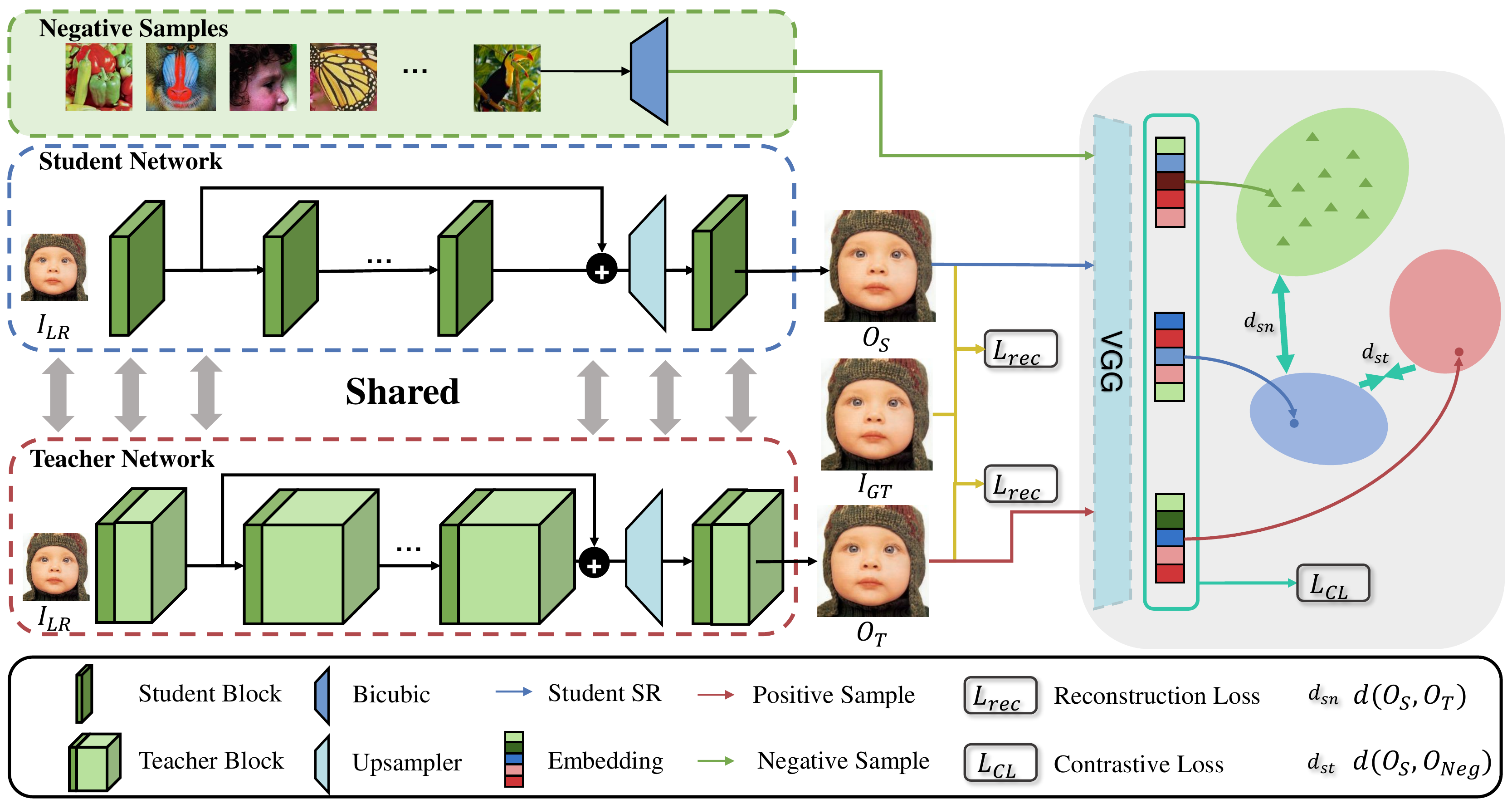}
\caption{The framework of our proposed CSD. Here we choose EDSR as our backbone for example. The darker green parts of the CSSR-Net (student) are shared from its teacher. The student and teacher separately produce a high-resolution image constrained by the reconstruction loss. The rich knowledge is constructed by contrastive loss and explicitly transferred from teacher to student, where the output $O_S$ is pulled to closer to the $O_T$ and pushed far away from the negative samples in the embedding space of VGG network.}
\label{Model} 
\end{figure*}

To answer the above question, we propose a \emph{contrastive self-distillation (CSD)} scheme to explicitly transfer the knowledge from teacher to CSSR-Net using contrastive loss (CL), which is inspired by contrastive learning \cite{oord2018representation,chen2020simple,he2020momentum}. 
As shown in Fig. \ref{Model}, SR model compression and acceleration is accomplished by CSSR-Net as a student. Meanwhile, CSD receives rich internal and external knowledge from reconstruction error and CL, respectively. 
There are two ``opposing forces'' on CL; One pulls the output of CSSR-Net closer to its teacher, the other one pushes the output of CSSR-Net farther away from the negative images in the latent feature space. CL constrains the output of CSSR-Net into the closed upper and lower bounds, which shows better quality images (see Fig. \ref{visualLossFig}). 

Our main contributions could be summarized as follows:
\begin{itemize}
    \item The proposed CSD scheme as a universal method can simultaneously compress and accelerate various SR networks which is also runtime friendly for practical use.
    \item Self-distillation is introduced to compress and accelerate SR models, while contrastive loss is proposed to further improve the performance of CSSR-Net by effective knowledge transferring. 
    \item Extensive experiments demonstrate the effectiveness of our CSD scheme . 
    For example, on Urban100, the compressed EDSR+ achieves $4\times$ compression rate and $1.77\times$ speedup, with only a minor loss of 0.13db PSNR and 0.0039 SSIM at the resolution scale of $\times$4.
\end{itemize}

\section{Related Work}
\subsection{Single Image Super-Resolution}
DCNNs based Super-resolution methods \cite{lim2017enhanced,zhang2018image,ahn2018fast} has shown impressive performance for SR in recent years.  
To improve the visual effect of the reconstructed images, the perceptual loss \cite{johnson2016perceptual} and the adversarial loss \cite{wang2018esrgan} were introduced. 
Both losses are effective as regularizers, 
but cannot prevent blurry areas as they are limited by the information of the ground truth images as an upper bound. 
Moreover, they require heavy parameters and computations to improve the quality of HR images, making them impractical to be employed on resource-limited embedded devices.

Several approaches have been proposed to reduce parameters for SR; they can be divided into recursive-based SR \cite{yang2018drfn,li2019srfbn}, quantization-based SR\cite{ma2019efficient,li2020pams} and compact architecture-based SR\cite{zhao2020efficient,hui2019lightweight}. Recently, AdderSR \cite{song2021addersr} utilize adder neural networks to avoid massive energy consumptions while GhostSR \cite{nie2021ghostsr} reduced parameters by generating ghost features. 
For recursive-based SR and quantization-based SR, they are difficult to accelerate SR models in practical applications, In contrast, Our
CSD scheme can simultaneously compress and accelerate various off-the-shelf SR models. %
Our CSD can also be integrated with compact architecture based SR models, which are orthogonal to our core contribution.

\subsection{Knowledge Distillation}
Knowledge Distillation (KD) \cite{hinton2015distilling} is a teacher-student framework that transfers information from a full teacher network to a compact student network. \cite{8478366}  achieve accelerating and compressing via low-rank decomposition with knowledge transfer.
Recently, self-distillation has been proposed~\cite{zhang2019your,ruiz2020distilled}, where knowledge is distilled within the network itself. \cite{liu2020metadistiller} proposes a stage-based self-distillation method termed as MetaDistiller by transferring the knowledge from deeper stages to earlier stages. However, few self-distillation methods are exploited for SR task. Different from MetaDistiller, our teacher and student have the same number of stages and knowledge is constructed explicitly via the contrastive loss rather than using a label generator to produce soft targets.

In line with our work, \cite{yu2018slimmable} proposed slimmable neural networks executing different widths on various classification models for dynamic resource adjustment. Training with different widths 
and switchable batch normalization is a class of implicit knowledge distillation which works well for image classification. However, it will leads to sub-optimal results \cite{yu2019universally} when directly applying these methods into SR task.
In contrast, we introduce a contrastive loss to explicitly transfer the knowledge from teacher to CSSR-Net via the closed upper and lower bound constraints,
which significantly improve the performance of CSSR-Net.

\subsection{Contrastive Learning}
Contrastive losses are widely used in self-supervised learning \cite{oord2018representation,he2020momentum,chen2020simple} and aim to pull the anchor close to positive points while pushing it away from negative
points in the representation space. 
Previous works \cite{chen2020simple,xu2020knowledge} have applied contrastive learning in high-level tasks. \cite{tian2020contrastive} presents a contrastive distillation framework to capture correlations of structured representations for image classification. Our work on the other hand uses contrastive learning to provide external knowledge with upper and lower bounds for SR tasks. Instead of using two separate networks, we adopt a self-distillation framework, facilitating dynamic resource adjustment by using different channel rates in the teacher model. 

Recently, \cite{park2020contrastive} employed contrastive learning to improve unpaired image-to-image translation quality. \cite{wu2021contrastive} proposed a contrastive regularization to improve the performance of various SOTA dehazing networks. Inspired by \cite{wu2021contrastive}, we introduce a contrastive loss for the SR task; contrastive learning is unexplored for SR and different from \cite{park2020contrastive} and \cite{wu2021contrastive}, we also introduce a new way to generate negative samples. 

\section{Method}

Our CSD contains two parts: CSSR-Net and contrastive loss (CL). First, we describe the CSSR-Net. Then, we present our CL to construct the upper and lower bound for CSSR-Net. Finally, the overall loss function of CSD scheme is presented and solved by a new optimization strategy.

\subsection{Channel-Splitting Super-Resolution Networks}
The channel-splitting super-resolution network (CSSR-Net) is a network constructed by splitting any CNN-based super-resolution network in the channel dimension. The CSSR-Net can be considered as a student network while the original network from which it is derived is a teacher network.  The pair can be used to construct a self-distillation framework and implicitly transferring the knowledge from the teacher to the student. CSSR-Net is entangled with the teacher network as it shares a portion of weights from the teacher, as illustrated in the left panel of Fig. \ref{Model}.

The width of the CSSR-Net is controlled by a manually set 
scale factor $r_w$ multiplied by the teacher's width uniformly at all layers. For example, $r_w = 0.5$, corresponds to the CSSR-Net retaining half of the width or number of channels as the teacher in all layers. 

Inspired by \cite{yu2018slimmable}, we can jointly minimize the reconstruction loss for both CSSR-Net and the teacher as:
\begin{equation}{
\label{eq1}
        \resizebox{.91\linewidth}{!}{
        $\begin{aligned}
         L_{Rec} &= \sum_i^N L_1(O_S^{(i)}, I_{GT}^{(i)})+\lambda_TL_1(O_T^{(i)}, I_{GT}^{(i)}\big), \\
            &=\sum_i^N L_1\big(f^S(I^{(i)},\theta_S), I_{GT}^{(i)}\big)+\lambda_TL_1\big(g^T(I^{(i)},\theta_T), I_{GT}^{(i)}\big),
        \end{aligned}$ }
}
\end{equation}
\noindent where $O_S^{(i)}=f^S(I^{(i)},\theta_S)$ and $O_T^{(i)}=g^T(I^{(i)},\theta_T)$ are the output of the CSSR-Net $f^S$ and the teacher network $g^T$ on the LR input $I^{(i)}$ with parameters $\theta_S$ and $\theta_T$ respectively. $\theta_S$ is shared from $\theta_T$ and satisfy $\theta_S\subset \theta_T$. $I_{GT}^{(i)}$ is the ground-truth HR image and $N$ is the number of training images. 

Directly minimizing Eq. \ref{eq1} via stochastic gradient descent (SGD) leads to sub-optimal results in SR task (See the results in Table \ref{tab:EDSR}).
As a result, CSSR-Net and the teacher network converge with worse results than the corresponding individual training. Moreover, the generated SR images have blurry parts, \emph{e.g.} the area on dense straight lines (see Fig. \ref{visualLossFig}). 
We speculate the following reason: implicit KD is not strong enough to provide insightful information 
by two independent loss terms. We expect to add explicit knowledge, which can provide richer internal and external knowledge.
Therefore, we introduce contrastive learning to explicitly construct a relationship between the student and the teacher, also providing the closed upper and lower bound to improve performances of both CSSR-Net and the teacher.
Upper bound is constructed to pull the output of CSSR-Net to teacher's, while lower bound is to constrain CSSR-Net's output to be far away from negative samples (\emph{e.g.} bicubic upsampled images).

\subsection{Contrastive Loss}
\label{method:ContraL}
Contrastive learning has improved representation learning by pulling anchors close to positive samples while pushing away negative samples~\cite{oord2018representation,park2020contrastive,chen2020simple,he2020momentum}. 
We propose a novel contrastive loss (CL) to explicitly represent 
knowledge for training both the teacher network and CSSR-Net. 
For contrastive learning, we need to consider two aspects: One is to construct
the ``positive'' and ``negative'' samples, the other
is to find the latent feature space to compare the samples. 

For the former, we construct the output $O_S^{(i)}$ of CSSR-Net and the output $O_T^{(i)}$ of its teacher as the anchor and the positive sample, respectively. 
%
More negative samples may 
better cover the undesired distribution. We thus sample $K$ images (other than the anchor) from the same mini-batch to $I^{(i)}$ as negative samples.
Then upsample them to the same resolution as $O_S^{(i)}$ via bicubic interpolation. Each of them is denoted by $O_{Neg}^{(k)},k=1,2\cdots,K$.
More detailed experimental results on numbers of negative samples are shown in experiments.

For the latent feature space, we use intermediate features of a pre-trained model $\phi$ (\emph{e.g.} VGG \cite{simonyan2014very}). 
%
Given positive and negative samples, we can construct the contrastive loss as:
\begin{equation}
\label{lcontra}
    L_{CL}=\sum_{i}^N\sum_{j}^M \lambda_j\frac{d\big(\phi_j(O_S^{(i)}),\phi_j(O_T^{(i)})\big)}{\sum_{k}^K d\big(\phi_j(O_S^{(i)}),\phi_j(O_{Neg}^{(k)})\big)},
\end{equation}
where $\phi_j, j=1,2,\cdots M$ is the intermediate features from the $j$-th layer of pre-trained model. 
$M$ is the number of total hidden layers.
$d(x,y)$ is the L1-distance loss between $x$ and $y$. $\lambda_j$ is the balancing weight for each layer. We do not update the parameters of pre-trained model $\phi$
during training. The contrastive loss is presented in Fig. \ref{Model} right panel; the loss introduces \emph{opposing forces} pulling the output of CSSR-Net $O_S^{(i)}$ to the output of its teacher $O_T^{(i)}$ and pushing $O_S^{(i)}$ to the negative samples $O_{Neg}^{(k)}$.
Note that our contrastive loss is different from InfoNCE \cite{oord2018representation}, which uses a dot product-based similarity. Instead of this similarity, our L1-distance loss achieves better performance (See experiments for more details). 
Additionally related to our CL is the perceptual loss \cite{johnson2016perceptual}, which minimizes the distance loss between the student and the ground-truth from multi-layer features of the pre-trained VGG.  This is an upper bound to constrain the student network.  Unlike the perceptual loss, however, 
we also adopt multiple negative samples as a lower bound to reduce the solution space and further improve the performance of CSSR-Net and its teacher.

\subsection{The Overall Loss and Its Solver}

\label{Train}
\paragraph{Overall Loss.} The overall loss of our CSD scheme is constructed by leveraging contrastive loss Eq. \ref{lcontra} into reconstruction loss Eq. \ref{eq1}, which can be formulated as:
\begin{equation}
\label{loss}
    L(\theta_S, \theta_T)=L_{Rec}+\lambda_{C}L_{CL},
\end{equation} 
where $\lambda_{C}$ is a hyper-parameter for balancing $L_{Rec}$ and $L_{CL}$. 
\paragraph{Solver.} Since our CSSR-Net and its teacher are entangled, we need to update both gradients from them. One naive solver is to update all parameters (\emph{i.e.}, $\theta_S$ and $\theta_T$) by directly minimize Eq. \ref{loss} based on SGD.
However, the teacher simply becomes weaker rather than maintain good performance. 

As a solution, we detach teacher's gradients from the contrastive loss and only update gradients from reconstruction loss. For student, we take normal gradient updating of Eq. \ref{loss}.
Pseudocode of CSD scheme is summarized in Algorithm \ref{alg:algorithm1}. 
\begin{lstlisting}[caption={Pseudocode of CSD in a PyTorch-like style}, label=alg:algorithm1, float=t]
# f_t: Pre-trained teacher network
# width_mult: the width of network f_s
# lt, lc: lambda_t, lambda_c in Eq. 1, Eq. 3
initialize()
for lr, hr in loader: # load a minibatch
    neg = bic(generate()) # negative samples
    o_s = f_s.forward(lr) # anchor
    o_t = f_t.forward(lr) # positive samples
    # reconstruction loss, Eq. 1
    loss = L1(o_s, hr) + lt * L1(o_t, hr)
    # No gradient to o_t from contrastive loss
    o_t.detach() 
    # contrastive loss, Eq. 2
    vgg_s, vgg_t, vgg_n = VGG19(o_s, o_t, neg)
    loss += lc * CL(vgg_s, vgg_t, vgg_n)
    loss.backward() # update
\end{lstlisting}

\section{Experiments}
\begin{figure*}[t]
\centering
    \begin{minipage}[b]{0.123\linewidth}
    {
    \includegraphics[width=1\linewidth]{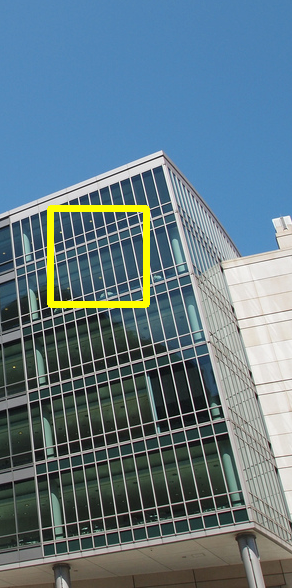}
    \begin{center}
        \tiny 096 from Urban100\\
        \quad
    \end{center}
    }
    \end{minipage}
    \begin{minipage}[b]{0.1\linewidth}
    {
    \includegraphics[width=1\linewidth]{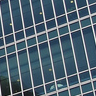}
    \begin{center}
        \tiny HR\\
        \tiny (PSNR / SSIM)
    \end{center}
    \includegraphics[width=1\linewidth]{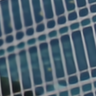}
    \begin{center}
        \tiny CSSR-Net\\
        \tiny (33.59 dB / 0.6357)
    \end{center}
    }
    \end{minipage}
    \begin{minipage}[b]{0.1\linewidth}
    {
    \includegraphics[width=1\linewidth]{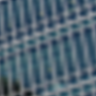}
    \begin{center}
        \tiny Bicubic\\
        \tiny (32.41 dB / 0.3138)
    \end{center}
    \includegraphics[width=1\linewidth]{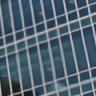}
    \begin{center}
        \tiny CSD (Ours) $0.25\times$\\
        \tiny (34.74 dB / 0.7888)
    \end{center}
    }
    \end{minipage}
    \begin{minipage}[b]{0.1\linewidth}
    {
    \includegraphics[width=1\linewidth]{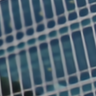}
    \begin{center}
        \tiny B-EDSR+ $0.25\times$\\
        \tiny (33.45 dB / 0.6048)
    \end{center}
    \includegraphics[width=1\linewidth]{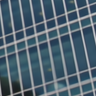}
    \begin{center}
        \tiny B-EDSR+ $1.0\times$\\
        \tiny (35.09 dB / 0.8263)
    \end{center}
    }
    \end{minipage}
    \begin{minipage}[b]{0.123\linewidth}
    {
    \includegraphics[width=1\linewidth]{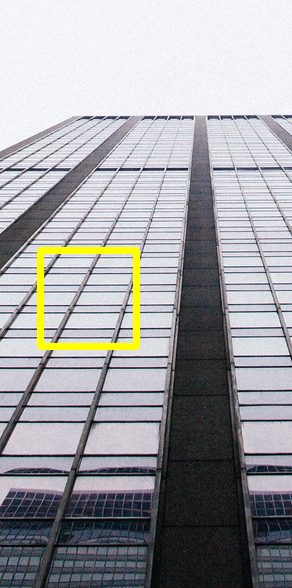}
    \begin{center}
        \tiny 0845 from DIV2K\\
        \quad
    \end{center}
    }
    \end{minipage}
    \begin{minipage}[b]{0.1\linewidth}
    {
    \includegraphics[width=1\linewidth]{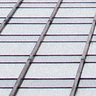}
    \begin{center}
        \tiny HR\\
        \tiny (PSNR / SSIM)
    \end{center}
    \includegraphics[width=1\linewidth]{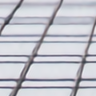}
    \begin{center}
        \tiny CSSR-Net\\
        \tiny (34.05 dB / 0.6801)
    \end{center}
    }
    \end{minipage}
    \begin{minipage}[b]{0.1\linewidth}
    {
    \includegraphics[width=1\linewidth]{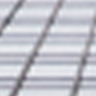}
    \begin{center}
        \tiny Bicubic\\
        \tiny (33.05 dB / 0.3632)
    \end{center}
    \includegraphics[width=1\linewidth]{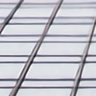}
    \begin{center}
        \tiny CSD (Ours) $0.25\times$\\
        \tiny (34.53 dB / 0.7742)
    \end{center}
    }
    \end{minipage}
    \begin{minipage}[b]{0.1\linewidth}
    {
    \includegraphics[width=1\linewidth]{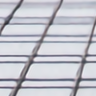}
    \begin{center}
        \tiny B-EDSR+ $0.25\times$\\
        \tiny (34.19 dB / 0.7027)
    \end{center}
    \includegraphics[width=1\linewidth]{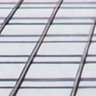}
    \begin{center}
        \tiny B-EDSR+ $1.0\times$\\
        \tiny (34.93 dB / 0.8080)
    \end{center}
    }
    \end{minipage}
    \caption{Qualitative Comparison on EDSR+ with $\times4$ SR. ``B-'' and CSSR-Net are baseline and EDSR+ 0.25$\times$ by J-T1, respectively.}
    \label{visualFig}
\end{figure*}

\subsection{Experimental Setups}
\paragraph{Implementation Details.}
Our CSD scheme is implemented by PyTorch 1.2.0 and MindSpore 1.2.0\cite{mindspore} with one NVIDIA TITAN RTX GPU. The models are trained with ADAM optimizer by setting $\beta_1 = 0.9, \beta_2 = 0.999$, and $\epsilon = 10^{-8}$. The batch size and total epochs are set to 16 and 300 epochs, respectively. The initial learning rate is $10^{-4}$ and decayed by 10$\times$ at every $2\times10^5$ iterations. %
For the latent features in Eq. \ref{lcontra}, We extract the features from the 1st, 3rd, 5th, 9th and 13th layers of the pre-trained VGG-19 while with the corresponding coefficients $\lambda_i, i=1,\cdots 5 $ to $\frac{1}{32}, \frac{1}{16}, \frac{1}{8}, \frac{1}{4}$ and 1, respectively. 
We set the hyper-parameters $\lambda_T$ and $ \lambda_C$ in Eq. \ref{lcontra} to 1 and 200, respectively. The scale factor for the width of CSSR-Net is default set to 0.25 if not specially stated. The input is randomly cropped into patches and augmented with random horizontal flip and $90^{\circ}$ rotation to produce the fixed $192\times192$ HR patches during training. 

\paragraph{Datasets.}
We train all SR models with 800 training images on DIV2K
and evaluate on the 100 validation images. 
We additionally test on four SR benchmarks: Set5\cite{bevilacqua2012low}, Set14\cite{zeyde2010single}, BSD100\cite{martin2001database} and Urban100\cite{huang2015single}.

\paragraph{Evaluation Metric.}
We calculate PSNR and SSIM on the Y channel, also evaluate the compression rate and real speedup on one NVIDIA TITANX RTX GPU.
\paragraph{Teacher Backbones.} 
To validate our CSD scheme, we choose models with different sizes and structures (EDSR+\cite{lim2017enhanced}, RCAN+\cite{zhang2018image} and CARN+\cite{ahn2018fast}\footnote{``+'' means self-ensemble technique \cite{lim2017enhanced}) is applied into the networks}) 
as teacher backbones, upon which we construct their corresponding CSSR-Nets as students.
%
Note that CARN+ is the most compact network with only 1.1M parameters, compared to RCAN+ (15.6M) and EDSR+ (43.1M). 
\paragraph{Baselines.} We select individually trained CSSR-Net and its teacher as our baseline. We also compare the proposed CSD with CSSR-Net, which only optimizes Eq. \ref{eq1} via joint training, denoted as J-T1.  

\subsection{Quantitative and Qualitative Results}
\subsubsection{Quantitative Results}
\begin{table*}[t]
\centering
\resizebox{.9\linewidth}{!}{
\small
\begin{tabular}{ccccccccccccc}
\toprule
\multirow{2}*{Model} & \multirow{2}*{Method} & \multirow{2}*{Scale} & \multicolumn{2}{c}{DIV2K} & \multicolumn{2}{c}{Set5} & \multicolumn{2}{c}{Set14} & \multicolumn{2}{c}{BSD100} & \multicolumn{2}{c}{Urban100}\\
&&&PSNR&SSIM&PSNR&SSIM&PSNR&SSIM&PSNR&SSIM&PSNR&SSIM\\
\midrule
\multirow{6}*{EDSR+ 0.25$\times$($S$)}
    &\multirow{2}*{Baseline}
        &$\times 2$&34.75&0.9463&38.07&0.9607& 33.68&0.9179 & 32.23&0.9004 & 32.28&0.9298\\
        &&$\times 4$&29.00&0.8379& 32.23&0.8957&28.63&0.7825&27.60&0.7366 & 26.07&0.7851\\
    \cline{2-13}
    &\multirow{2}*{J-T1}
        &$\times 2$&34.73&0.9461&38.09&0.9608&33.67&0.9179&32.22&0.9002&32.26&0.9296\\
        &&$\times 4$&29.00&0.8379&32.22&0.8958&28.63&0.7827&27.60&0.7369&26.06&0.7852\\
    \cline{2-13}
    &\multirow{2}*{CSD}
        &$\times 2$&\textbf{34.85}&\textbf{0.9473} &\textbf{38.19}&\textbf{0.9612} & \textbf{33.85}&\textbf{0.9199} & \textbf{32.29}&\textbf{0.9012}& \textbf{32.55}&\textbf{0.9327} \\
        &&$\times 4$& \textbf{29.13}&\textbf{0.8416} & \textbf{32.34}&\textbf{0.8974}  & \textbf{28.72}&\textbf{0.7856} & \textbf{27.68}&\textbf{0.7396} & \textbf{26.34}&\textbf{0.7948}\\
\midrule
\multirow{6}*{EDSR+ 1.0$\times$($T$)}
    &\multirow{2}*{Baseline}
        &$\times 2$&35.12&\textbf{0.9699} &38.20&0.9606  & 34.02&0.9204 & 32.37&0.9018 & 33.10&0.9363\\
        &&$\times 4$&29.33&0.8456&\textbf{32.60}&0.8998&28.90&0.7892&27.76&0.7428&26.76&0.8061\\
    \cline{2-13}
    &\multirow{2}*{J-T1}
        &$\times 2$&35.11&0.9489&38.29&0.9615&34.10&0.9216&32.40&0.9024&32.13&0.9371\\
        &&$\times 4$&29.32&0.8453&32.59&0.8997&28.89&0.7889&27.76&0.7424&26.73&0.8055\\
    \cline{2-13}
    &\multirow{2}*{CSD}
        &$\times 2$&\textbf{35.15}&0.9402&\textbf{38.31}&\textbf{0.9617}&\textbf{34.07}&\textbf{0.9216}&\textbf{32.41}&\textbf{0.9026}&\textbf{33.17}&\textbf{0.9375}\\
        &&$\times 4$&\textbf{29.34}&\textbf{0.8458}&\textbf{32.60}&\textbf{0.9000}&\textbf{28.91}&\textbf{0.7893}&\textbf{27.77}&\textbf{0.7430}&\textbf{26.79}&\textbf{0.8063}\\
\bottomrule
\end{tabular}}

\caption{Results on EDSR+. $S$ and $T$ respectively indicate EDSR+ 0.25$\times$ and EDSR+ 1.0$\times$, where the number 0.25 or 1.0 is the width scale. EDSR+ 0.25$\times$ has 2.7M parameters, which achieves 16$\times$ compression rate compared to $T$. }
\label{tab:EDSR}
\end{table*}

We first compress EDSR+, as shown in Table \ref{tab:EDSR}. EDSR+ 1.0$\times$ denotes the original EDSR+ model as a teacher, while EDSR+ 0.25$\times$ presents a student with 2.7M parameters by setting $r_w = 0.25$. Compared to the teacher, EDSR+ 0.25$\times$ achieves 16.0$\times$ compression rate.
We can observe that (1) Joint training (J-T1) achieves the consistent PSNR/SSIM with baseline (\emph{i.e.} individual training) both in student and teacher at the same SR scale. For dynamic inference, the baseline needs to store two individual models with different parameters 
, while J-T1 
only requires the parameter of the teacher; (2) Compared to baseline and J-T1, CSD achieves the best performance both in student and teacher at the same SR scale, except the SSIM metric of teacher at SR scale of 2 on DIV2K. For example, on Urban100, EDSR+ 0.25$\times$ using CSD achieves the PSNR gains over baseline by 0.26dB and 0.24dB at the SR scale of 2 and 4, respectively; (3) Using CSD, our EDSR+ 0.25$\times$ model achieves 16$\times$ compression rate, only with the loss of 0.2db PSNR at the SR scale of 4 (\emph{i.e.} 29.13dB \emph{vs.} 29.33dB in baseline) on DIV2K. 

\begin{figure}[t]
\centering  
\subfigure[PSNR-parameter]{
\label{psnr}
\includegraphics[width=0.48\linewidth, height=0.9in]{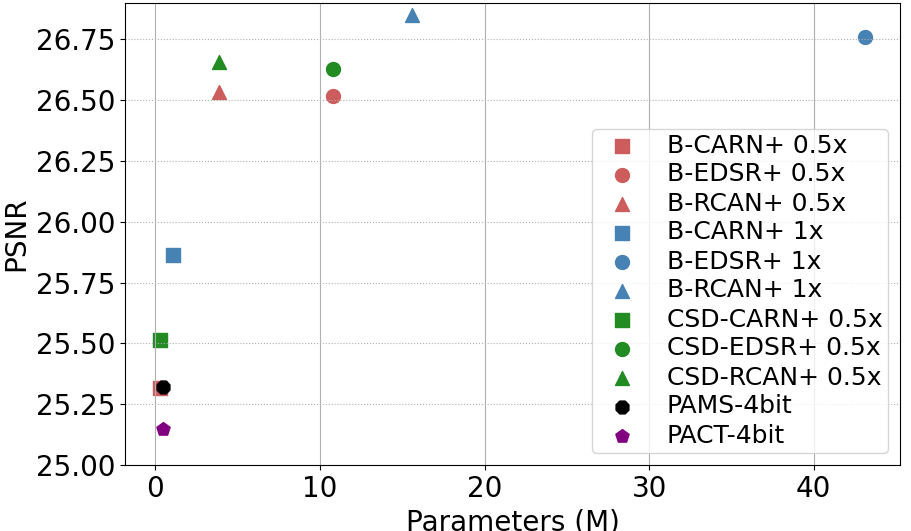}}
\subfigure[PSNR-Speed]{
\label{speed}
\includegraphics[width=0.48\linewidth, height=0.9in]{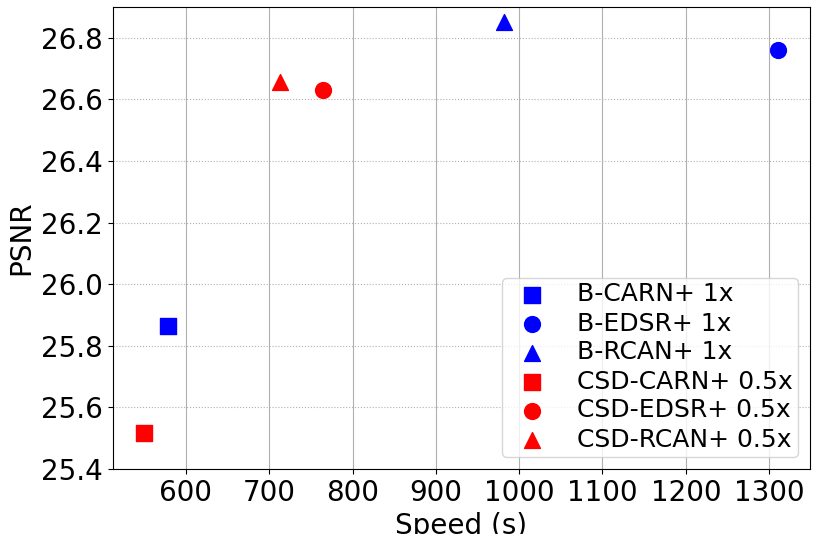}}
\caption{Results of EDSR+, RCAN+, CARN+ compression on Urban100 $\times 4$, GPU speedup rate and PSNR of their corresponding students (width=0.5).}
\label{psnr_rate}
\end{figure}

We further compress RCAN+ and CARN+, which are more compact. Additionally, we set $r_w = 0.5$. As shown in Fig. 4, our CSD achieves higher PSNR at the SR scale of 4 on Urban100, compared to baseline. 
For example, on Urban100 $\times 4$, our CSD-RCAN+ 0.5$\times$ achieves 0.12dB PSNR gains over B-RCAN+ 0.5$\times$. We also found that CARN+ compression is relatively difficult, which is due to the smallest number of redundant parameters. However, our CSD still achieves higher PSNR, compared to the baseline based on individual training. Moreover, compared to SOTA SR quantization methods (\emph{i.e.} PAMS-4bit \cite{li2020pams} and PACT-4bit \cite{choi2018pact}), our CSD achieves 0.19dB and 0.37dB PSNR gains over PAMS and PACT, while with the smallest parameter number of 0.3M (vs. 0.48M\footnote{This number has been converted to 32-bit float number.} both in PAMS and PACT).
For fair comparison to evaluate the speedup, we report the total inference time across Urban100 dataset with SR scalar of 4 (See Fig. \ref{speed}). 
Our CSD achieves $3.9\times$ compression rate and $1.38\times$ GPU speedup rate only with 0.2dB PSNR loss, compared to the original model.   

\subsubsection{Qualitative Results}

As shown in Fig. \ref{visualFig}, we compare our CSD with other training methods on the quality of the enhanced images at $4\times$ SR scale on Urban100, BSD100 and DIV2K.
For simplicity, we select the student model EDSR+ $0.25\times$ for visual comparison.
We can see that both outputs images based on individual training (baseline) and joint training (CSSR-Net) are blurry, especially in the areas where straight lines are very dense. 
In contrast, our CSD can effectively alleviate the blurry problem, which means that high frequency information is effectively recovered. More examples are presented in Supplementary.

\subsection{Ablation Study}
To demonstrate the effectiveness of the proposed CSD scheme, we conduct ablation study to analyze the effect of contrastive loss, S-T distillation, updating strategies and the number of negative samples. EDSR+ is selected as our backbone in ablation study.

\paragraph{Effect of InfoNCE Loss, Perceptual Loss and CL.} InfoNCE loss \cite{oord2018representation,he2020momentum} has been widely used in contrastive learning. It uses the dot-product operation to measure the distance between two vectors. Instead of InfoNCE loss, we use contrastive loss in Eq. \ref{lcontra} based on L1-distance measurement. As shown in Tab. \ref{table3}, our CL achieves the best results, compared to InfoNCE and J-T with perceptual loss. For example, Our CSD based on L1-distance achieves 0.0014 and 0.0041 SSIM gains over InfoNCE based CSD on DIV2K and Urban100, respectively.

\label{dAB}
\begin{table}[t]
\centering
\resizebox{.85\linewidth}{!}{
\small
\begin{tabular}{lcccc}
\toprule
\multirow{2}*{Methods}
    & \multicolumn{2}{c}{DIV2K} &\multicolumn{2}{c}{Urban100}\\
    &PSNR&SSIM&PSNR&SSIM\\
\midrule
Baseline  & 29.00&0.8379 & 26.07&0.7851\\
CSD (InfoNCE) & 29.10&0.8402 & 26.24&0.7907\\
J-T (W/ Perceptual Loss) & 28.72 & 0.8317 & 25.72&0.7757 \\
\hline
W/O T (GT Pos.) & 29.11&0.8406 & 26.30&0.7924\\
CSD (GT Pos.) & 28.87 & 0.8356 & 26.00& 0.7844 \\
T-S Separate & 29.03&0.8406 & 26.15&0.7894\\
\hline
CSD (Ours)  & {\textbf{29.13}}&\textbf{0.8416} & {\textbf{26.34}}&\textbf{0.7948}\\
\bottomrule
\end{tabular}}

\caption{Comparison of different contrastive losses and S-T distillations on EDSR+ 0.25$\times$ model on Urban100 $\times$4.}
\label{table3}
\end{table}
\paragraph{Effect of S-T Distillation}
\label{ABArc}
We further evaluate the effectiveness of our CSSR-Net. We consider three S-T distillation strategies: (1) W/O T (GT Pos.) only use ground-truth HR images as positive samples, which remove the teacher's branch; (2) CSD (GT Pos.) uses ground-truth HR images as postive samples in CSD scheme; (3) T-S Separate where the student does not share weights with the teacher. 

As shown in Table \ref{table3}, our CSSR-Net enables self-distillation, which receives richer information from its entangled teacher. 
It is worth noticing that the performance of CSD is better than CSD (GT Pos.). We speculate that HR images provide a stronger upper bound which is more difficult for the limited capacity S to fully exploit. This is consistent with findings in \cite{cho2019efficacy}, where better teachers do not necessarily help to train better students. 

\paragraph{Effect of Updating Strategy.}
\label{ContraAB}
\begin{figure}[t]
\centering 
\includegraphics[width=0.95\linewidth, height=1in]{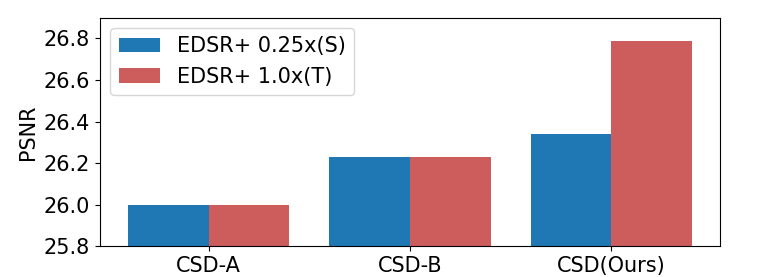}
\caption{Comparison of updating strategies on Urban100 $\times 4$. }
\label{ABlossFig} 
\end{figure}
Fig. \ref{ABlossFig} presents a comparison of different updating strategies on CL. CSD-A removes the reconstruction loss for $T$ in Eq. \ref{loss}; CSD-B directly minimize Eq. \ref{loss} by both updating weights of $S$ and $T$ from the gradient of CL; CSD detaches $T$'s gradient from CL. 
Obviously, our CSD achieves significantly higher performance, compared to CSD-A and CSD-B. 
In CSD-A, $T$ collapses quickly in the early stage during training and is unable to provide valid information to $S$, resulting in much worse performance. In CSD-B, $T$ tends to learn the student such that $T$ become weaker during training. 

\paragraph{Effect of the Number of Negative Samples.}
\label{ABNeg}

\begin{figure}[t]
\centering 
\includegraphics[width=0.95\linewidth, height=1in]{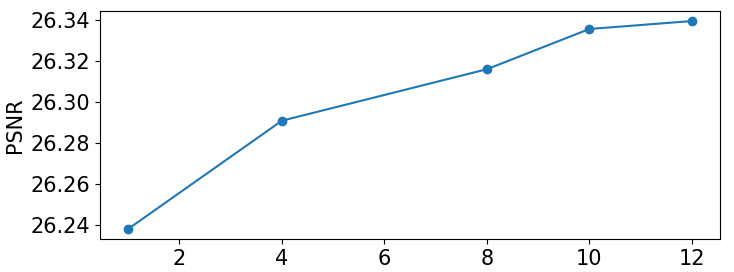}
\caption{Comparison of numbers of negative samples. We evaluate PSNR scores of CSD-EDSR+ $0.25\times$ on Urban100 $\times4$. }
\label{ABnegFig} 
\end{figure}
Finally, we explore the effect of the number of negative samples. As shown in Fig. \ref{ABnegFig}, adding more negative samples achieves better performance.
However, training memory and time grows as the number of negative samples increases. For the performance-efficiency trade-off, we choose the number of 10 negative samples in our all experiments.

\section{Conclusion}
In this paper, we propose a novel CSD scheme for simultaneously compressing and accelerating SR models, which consists of CSSR-Net and contrastive loss. Constructed from the target (teacher) network, CSSR-Net shares the part weights, directly compressing the target network. To mine for rich knowledge from the teacher, a novel contrastive loss is proposed for explicit knowledge transfer, which ensures that the output of CSSR-Net is pulled closer to the teacher's and pushed far away from the blurry images. We have comprehensively evaluated the performance of CSD scheme for compressing and accelerating various SR models on standard benchmark datasets with superior performance. We believe the proposed CSD scheme can be generalized to other low-level vision tasks (\emph{e.g.} dehazing, denoising and debluring), which will be explored in future work. 

\section*{Acknowledgements}
This work is sponsored by National Natural Science
Foundation of China (61972157, 61772524, 61876161,
61902129), Natural Science Foundation of Shanghai
(20ZR1417700), CAAI-Huawei MindSpore Open Fund,
National Key Research and Development Program of
China (2019YFC1521104), Zhejiang Lab (2019KD0AC02, 2020NB0AB01), the Fundamental Research Funds for the Central Universities, Shanghai Sailing Program (21YF1411200). We thank MindSpore\cite{mindspore} for their partial support.

\small
\bibliographystyle{named}
\bibliography{ijcai21}

\end{document}